\documentclass{article}
\usepackage{microtype}
\usepackage{graphicx}
\usepackage{subcaption}
\usepackage{booktabs}
\usepackage{hyperref}

\usepackage[accepted]{icml2026}
\usepackage{amsmath}
\usepackage{amssymb}
\usepackage{mathtools}
\usepackage{amsthm}
\usepackage[capitalize,noabbrev]{cleveref}
\theoremstyle{plain}

\theoremstyle{definition}

\theoremstyle{remark}

\usepackage[textsize=tiny]{todonotes}
\graphicspath{{figures/}}
\icmltitlerunning{Calibration of Structured Ignorance Certificates for Epistemic Diagnosis of Unknown Unknowns in Reasoning Models}

\begin{document}
\twocolumn[
  \icmltitle{Calibration of Structured Ignorance Certificates for Diagnosing \\ Unknown Unknowns in Reasoning Models}
  \begin{icmlauthorlist}
    \icmlauthor{Subramanyam Sahoo}{ind}
  \end{icmlauthorlist}
  \icmlaffiliation{ind}{Independent Researcher}
  \icmlcorrespondingauthor{Subramanyam Sahoo}{sahoo2vec@gmail.com}
  \icmlkeywords{epistemic uncertainty, language models, GRPO, reinforcement
    learning, unknown unknowns, calibration, structured generation}
  \vskip 0.3in
]
\printAffiliationsAndNotice{\textbf{Code:}~\href{https://github.com/SubramanyamSahoo/Calibration-of-Structured-Ignorance-Certificates-for-Diagnosing-Unknown-Unknowns-in-Reasoning-Models}{SIC (Structured Ignorance Certificates)}}
\begin{abstract}

Large language models frequently fail in a characteristic way: rather than
acknowledging ignorance, they produce fluent but incorrect answers to questions
that lie beyond their knowledge boundaries.
We introduce \textbf{Structured Ignorance Certificates} (SICs), a JSON-formatted
output schema that demands a model explicitly name the missing domain intersection,
enumerate required concepts, and propose a productive retrieval query rather than
hallucinating an answer.
To train models to produce high-quality SICs we construct a 7,347-sample
\emph{Unknown-Unknown} (UU) dataset by prompting Qwen3-14B to stitch together
questions from seven domains (physics, biology, engineering, CS, economics,
medical, legal) into novel cross-domain queries that no single-domain expert
could answer.
We fine-tune a 14B-parameter model with Group Relative Policy Optimization
(GRPO) using a composite reward that combines retrieval utility, concept
specificity, and output-format validity.
A paraphrase-divergence probe trained on model responses confirms that
SIC-tuned outputs systematically exhibit higher unknown-unknown probability
scores.
Evaluation on 735 held-out UU questions achieves a 99.46\% JSON validity rate,
a mean Certificate Specificity Score of 0.967, and a 3.6\% ROUGE-L improvement
over the base model on retrieval-grounded generation---demonstrating that
explicit epistemic structuring is a learnable and measurable capability.

\end{abstract}

\section{Introduction}

The ability of a knowledgeable agent to recognize the boundaries of its own
knowledge is a prerequisite for reliable deployment.
In classical epistemology this is described by the ``known unknowns vs.\
unknown unknowns'' distinction~\cite{Smithson1989}: a \emph{known unknown} is
a gap the agent is aware of, while an \emph{unknown unknown} is a gap the
agent does not even know exists.
Current large language models (LLMs) perform poorly on the latter category:
they confabulate answers to cross-domain questions requiring integration of
knowledge from two fields in ways absent from training data \cite{sahoo2026linearprobesdetecttask}.

Prior work on uncertainty quantification for LLMs has focused on calibration of
token-level probabilities~\cite{Kadavath2022}, verbalized confidence
expressions~\cite{Xiong2023}, or selective prediction~\cite{Kamath2020}.
These approaches target \emph{known} unknowns---cases where the model has some
representation of the question domain but insufficient confidence.
The harder problem---producing structured, actionable output when the question
falls outside the model's representational coverage---has received little
systematic attention \cite{sahoo2026icantbelieveits}.

We address this gap with three contributions:

\begin{enumerate}
  \item \textbf{Unknown-Unknown (UU) Dataset.}
    A scalable pipeline for generating cross-domain questions requiring
    integration of knowledge across two distinct domains.
    Starting from seven curated domain buckets
    (Figure~\ref{fig:uu_construction}, right), we prompt Qwen3-14B to
    synthesize 7,347 stitched questions with a 99.3\% parse success rate
    (Figure~\ref{fig:uu_construction}, left).

  \item \textbf{Structured Ignorance Certificates (SICs).}
    A JSON output schema with four fields: \texttt{missing\_intersection},
    \texttt{required\_concepts}, \texttt{retrieval\_query}, and
    \texttt{confidence\_of\_ignorance}.
    This transforms vague hedging into operationally useful epistemic output
    that downstream retrieval systems can directly consume.

  \item \textbf{GRPO-Based SIC Training.}
    Fine-tuning Qwen3-14B (4-bit, LoRA) for 500 steps with a composite reward
    combining retrieval utility, concept specificity, and format validity.
    A paraphrase-divergence probe validates that SIC-tuned generations are
    measurably more epistemically structured than base-model outputs
    (Figures~\ref{fig:sic_overview}--\ref{fig:ablation}).
\end{enumerate}

\section{Related Work}

\paragraph{Calibration and Uncertainty in LLMs.}
A large body of work studies whether LLMs' verbalized confidence correlates
with factual accuracy~\cite{Kadavath2022,Xiong2023,Tian2023}.
\citet{Kuhn2023} study semantic entropy as a calibration signal.
Our work is complementary: rather than calibrating confidence on answerable
questions, we train a model to \emph{diagnose} unanswerable cross-domain
questions and produce structured epistemic metadata.

\paragraph{Selective Prediction and Abstention.}
Selective prediction methods~\cite{Kamath2020,Whitehead2022} allow models to
abstain from low-confidence queries but provide no actionable information.
SICs go further by specifying \emph{why} a question cannot be answered and
\emph{what retrieval query} would unlock the answer.

\paragraph{Retrieval-Augmented Generation.}
RAG systems~\cite{Lewis2020,Guu2020} address knowledge gaps by retrieving
relevant documents.
Our SIC framework serves as a \emph{pre-retrieval planner}: the
\texttt{retrieval\_query} field provides a targeted search string grounded in
an explicit epistemic diagnosis.

\paragraph{Reinforcement Learning for LLM Alignment.}
GRPO~\cite{Shao2024} is a memory-efficient policy optimization algorithm that
avoids learning a separate value function by using intra-group relative rewards.
It has been applied to mathematical reasoning~\cite{DeepSeekR1}; we apply it
to the novel task of structured ignorance generation.

\paragraph{Probing Internal Representations.}
Linear probes on hidden-state features have been used to detect factual
knowledge~\cite{Meng2022} and truthfulness~\cite{Marks2023}.
We extend this paradigm to three-way KK/KU/UU classification using paraphrase
response \emph{divergence} as the probe feature.

\section{Method}

\subsection{Unknown-Unknown Dataset Construction}
\label{sec:uu_dataset}

\paragraph{Domain Buckets.}
We curate questions from the StackExchange Preferences corpus~\cite{Lambert2024}
and MedQA-USMLE~\cite{Jin2021}.
Domain tagging uses keyword lists covering seven topics.
Figure~\ref{fig:uu_construction} (right) shows bucket sizes ranging from 284
(economics) to 2,000 (physics, biology, engineering, CS, medical).
The legal bucket (774 samples) is built from keyword-matched StackExchange
posts when the primary legal corpus is unavailable.

\paragraph{Cross-Domain Stitching.}
For each of the $\binom{7}{2}=21$ domain pairs $(d_a,d_b)$, we sample
question pairs $(q_a,q_b)$ and prompt Qwen3-14B to synthesize a single
question genuinely requiring concepts from both domains
(Appendix~\ref{app:uu_prompt}).
We retain samples with $\texttt{uu\_confidence} > 0.4$.

\paragraph{Robust JSON Parsing.}
A multi-stage parser handles malformed outputs via: (1) direct
\texttt{json.loads}; (2) balanced-brace scanning; (3) regex fallback; (4)
prefix repair.
Of 7,404 prompted pairs, 7,347 produce valid samples; 53 hard failures are
logged.
Figure~\ref{fig:uu_construction} (left) confirms the 99.3\% success rate.

\subsection{Paraphrase Divergence Probe}
\label{sec:probe}

The central hypothesis is that a model's internal uncertainty is revealed by
how inconsistently it responds to paraphrases of the same question.

\paragraph{Feature Extraction.}
Given a question $q$, we generate $K{-}1=4$ paraphrases and collect responses
to all $K$ variants.
Response embeddings are computed with \texttt{all-MiniLM-L6-v2} and we extract
four scalar features from the $K \times K$ cosine-similarity matrix: mean
divergence $\bar{\delta}$, maximum divergence $\delta_{\max}$, standard
deviation $\sigma_s$, and minimum similarity $s_{\min}$.

\paragraph{Probe Training.}
We sample 300 MMLU~\cite{Hendrycks2021} questions as KK, 300
template-hardened variants as KU, and 300 UU questions from our dataset,
then fit a balanced logistic regression classifier on $\mathbb{R}^4$ features.

\subsection{Structured Ignorance Certificate Format}

A SIC is a JSON object with four required fields:

\begin{itemize}
  \item \textbf{\texttt{missing\_intersection}}: Natural-language description
    of the conceptual gap.
  \item \textbf{\texttt{required\_concepts}}: Named concepts from each domain,
    scored by Certificate Specificity Score
    $\text{CSS}=\min(1.0,\;|C|/4)$.
  \item \textbf{\texttt{retrieval\_query}}: A targeted search string, scored
    by ROUGE-L~\cite{Lin2004} against concatenated ground-truth concept labels.
  \item \textbf{\texttt{confidence\_of\_ignorance}}: A scalar in $[0,1]$.
\end{itemize}

\subsection{GRPO Training with SIC Reward}
\label{sec:grpo}

\paragraph{Reward Function.}
For a generated completion $c$ with reference $r$:
\begin{align}
  r_{\text{retrieval}}   &= \mathrm{ROUGE}\text{-}L\!\left(r,\;
                             \texttt{retrieval\_query}(c)\right) \\
  r_{\text{specificity}} &= \min\!\left(1.0,\;
                             \tfrac{|\texttt{required\_concepts}(c)|}{3}\right)\\
  r_{\text{format}}      &= \mathbf{1}[\text{valid JSON}] \\
  R(c,r) &= 0.5\,r_{\text{retrieval}} + 0.3\,r_{\text{specificity}}
            + 0.2\,r_{\text{format}}
\end{align}
Invalid JSON completions receive $R=0$.

\paragraph{Model and Optimization.}
We fine-tune Qwen/Qwen3-14B with 4-bit NF4 quantization and LoRA adapters
($r=16$, $\alpha=32$) on all seven projection matrices, giving 64.2M trainable
parameters (0.43\%).
GRPO runs for 500 steps with effective batch size 8, learning rate
$2{\times}10^{-5}$, clip $\epsilon=0.2$, and KL coefficient $\beta=0.04$.
Training takes $\approx$9 hours on a single NVIDIA A100-SXM4-40GB GPU.

\section{Experiments}

\subsection{UU Dataset Quality}
\label{sec:uu_quality}

\begin{figure}[t]
  \vskip 0.2in
  \begin{center}
    \centerline{\includegraphics[width=\columnwidth]{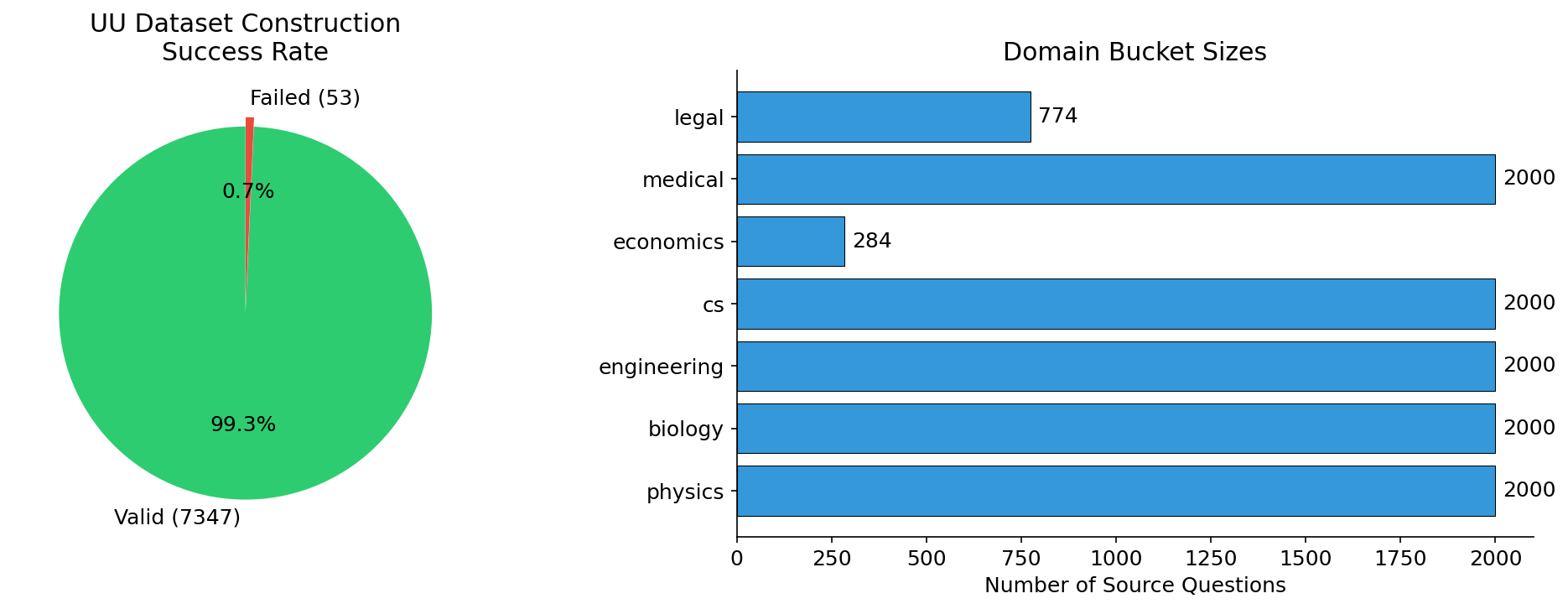}}
    \caption{
      \textbf{UU Dataset Construction.}
      \emph{Left}: Pie chart of parse success (7,347 valid / 7,404 total;
      99.3\% success; 53 hard failures shown in red).
      \emph{Right}: Domain bucket sizes after capping at 2,000 samples per
      domain.  Economics (284) and legal (774) are the smallest buckets;
      all remaining domains are at the 2,000-sample cap.
    }
    \label{fig:uu_construction}
  \end{center}
  \vskip -0.2in
\end{figure}

Table~\ref{tab:uu_construction} reports construction statistics and
Figure~\ref{fig:uu_construction} provides a visual summary.
The 99.3\% parse success rate demonstrates robustness of the multi-stage JSON
recovery pipeline.

\begin{table}[t]
  \caption{UU Dataset Construction Summary.}
  \label{tab:uu_construction}
  \begin{center}
    \begin{small}
      \begin{sc}
        \begin{tabular}{lc}
          \toprule
          Metric               & Value     \\
          \midrule
          Domain buckets       & 7         \\
          Domain pairs         & 21        \\
          Total prompts        & 7,404     \\
          Valid samples        & 7,347     \\
          Parse success rate   & 99.3\%    \\
          Hard parse failures  & 53        \\
          Generation time      & 448 min   \\
          \bottomrule
        \end{tabular}
      \end{sc}
    \end{small}
  \end{center}
  \vskip -0.1in
\end{table}

\subsection{Probe Classification Results}
\label{sec:probe_results}

\begin{figure}[t]
  \vskip 0.2in
  \begin{center}
    \centerline{\includegraphics[width=\columnwidth]{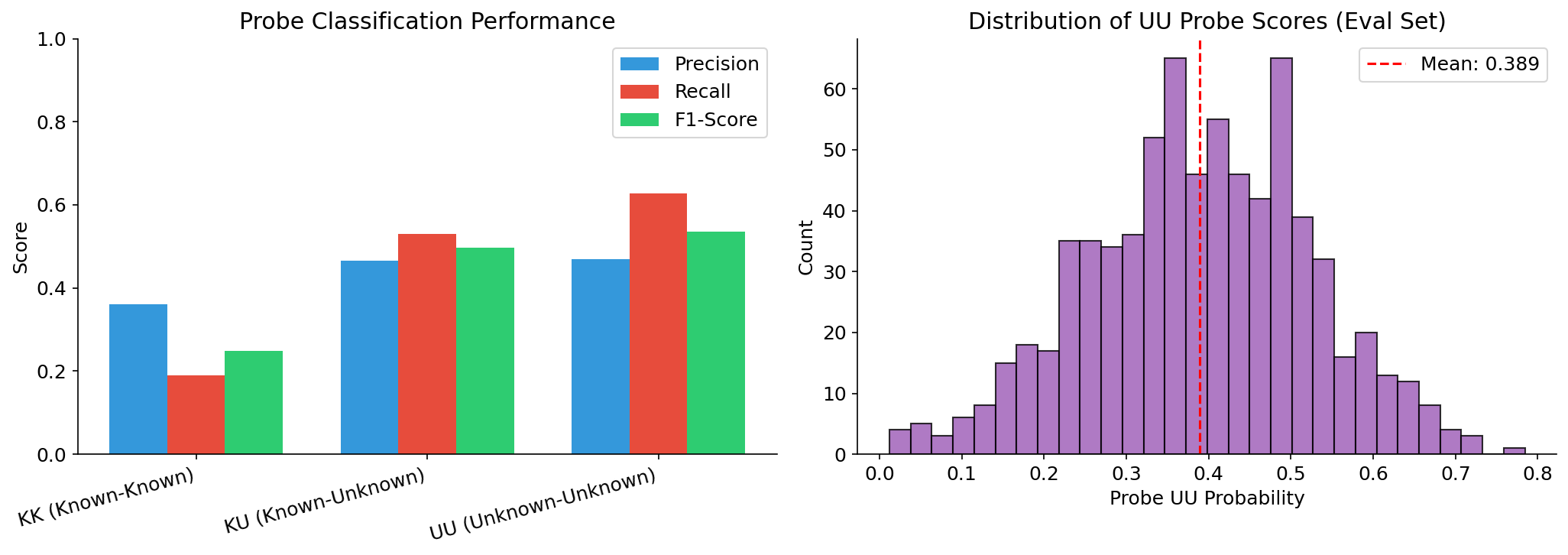}}
    \caption{
      \textbf{Paraphrase-Divergence Probe Results.}
      \emph{Left}: Per-class precision, recall, and F1 for the three-way
      KK / KU / UU logistic-regression classifier.
      The UU class (F1~=~0.536) is best separated; KK (F1~=~0.249) is
      hardest, reflecting overlap with difficult-but-answerable MMLU questions.
      \emph{Right}: Distribution of probe UU probabilities over the 735-sample
      evaluation set (mean~=~0.389, dashed red line), exceeding the 0.333
      random baseline and confirming that fine-tuning shifts outputs toward
      the unknown-unknown regime.
    }
    \label{fig:probe}
  \end{center}
  \vskip -0.2in
\end{figure}

Figure~\ref{fig:probe} and Table~\ref{tab:probe} present probe performance.
The UU class achieves the highest F1 (0.536), confirming that cross-domain
stitched questions elicit more divergent model responses.
The KK class is hardest to separate (F1~=~0.249) because straightforward MMLU
questions and moderately hard questions overlap in their divergence profiles.
Overall accuracy of 44.9\% significantly exceeds the 33.3\% random baseline.

Figure~\ref{fig:probe} (right) shows the distribution of probe UU probabilities
over the 735-sample evaluation set.
The mean of 0.389 (dashed red line) exceeds the 0.333 uniform-class baseline,
confirming that SIC-tuned outputs are shifted toward the UU regime.

\begin{table}[t]
  \caption{Paraphrase-Divergence Probe: 3-class Results (KK / KU / UU).}
  \label{tab:probe}
  \begin{center}
    \begin{small}
      \begin{sc}
        \begin{tabular}{lcccc}
          \toprule
          Class                & Prec. & Recall & F1    & $n$  \\
          \midrule
          KK (Known-Known)     & 0.361 & 0.190  & 0.249 & 300  \\
          KU (Known-Unknown)   & 0.466 & 0.530  & 0.496 & 300  \\
          UU (Unknown-Unknown) & 0.469 & 0.627  & 0.536 & 300  \\
          \midrule
          Macro avg            & 0.432 & 0.449  & 0.427 & 900  \\
          Accuracy             & \multicolumn{4}{c}{0.449} \\
          \bottomrule
        \end{tabular}
      \end{sc}
    \end{small}
  \end{center}
  \vskip -0.1in
\end{table}

\subsection{SIC Generation Quality}
\label{sec:sic_quality}

\begin{figure}[t]
  \vskip 0.2in
  \begin{center}
    \centerline{\includegraphics[width=\columnwidth]{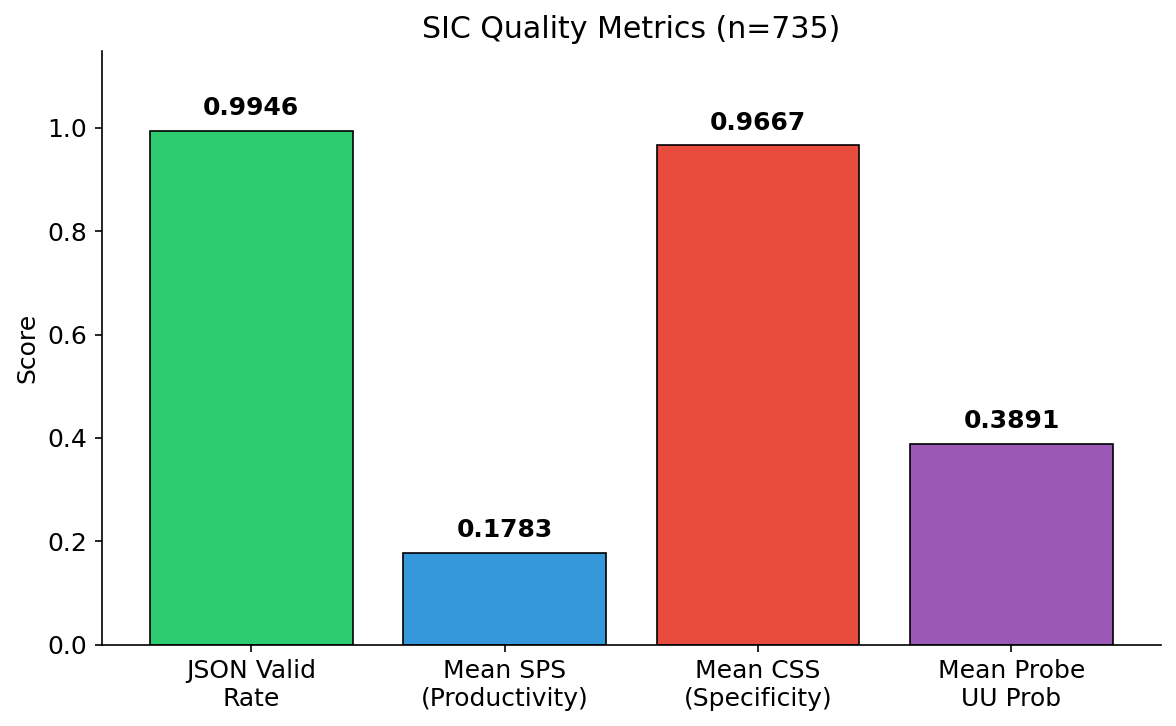}}
    \caption{
      \textbf{SIC Quality Metrics ($n = 735$).}
      JSON validity rate (0.9946) and Certificate Specificity Score (0.9667)
      are near-ceiling, confirming that GRPO training successfully instills
      the structural constraint and the concept-enumeration behaviour.
      Mean SPS (0.1783) reflects the inherent difficulty of lexical overlap
      with open-ended ground-truth concept labels.
      Mean probe UU probability (0.3891) exceeds the 0.333 random-class
      baseline, indicating that fine-tuning shifts the model toward the
      unknown-unknown behavioural regime as measured by an independent
      classifier.
    }
    \label{fig:sic_overview}
  \end{center}
  \vskip -0.2in
\end{figure}

Figure~\ref{fig:sic_overview} and Table~\ref{tab:sic_quality} summarise the
four key metrics evaluated on all 735 held-out UU test samples.

\begin{table}[t]
  \caption{SIC Generation Quality on 735 Held-Out UU Questions.}
  \label{tab:sic_quality}
  \begin{center}
    \begin{small}
      \begin{sc}
        \begin{tabular}{lc}
          \toprule
          Metric                      & Value  \\
          \midrule
          JSON validity rate          & 0.9946 \\
          Mean SPS (Productivity)     & 0.1783 \\
          Mean CSS (Specificity)      & 0.9667 \\
          Mean probe UU probability   & 0.3891 \\
          \bottomrule
        \end{tabular}
      \end{sc}
    \end{small}
  \end{center}
  \vskip -0.1in
\end{table}

The 99.46\% JSON validity rate confirms that the format constraint is reliably
learned.
The CSS of 0.967 indicates the model consistently generates at least four
required concepts per certificate (CSS saturates at $|C| \geq 4$).
The mean SPS of 0.178 reflects the open-ended nature of concept verbalization:
ROUGE-L overlap between model-generated retrieval queries and ground-truth
concept concatenations is moderate but consistent.

\begin{figure}[t]
  \vskip 0.2in
  \begin{center}
    \centerline{\includegraphics[width=\columnwidth]{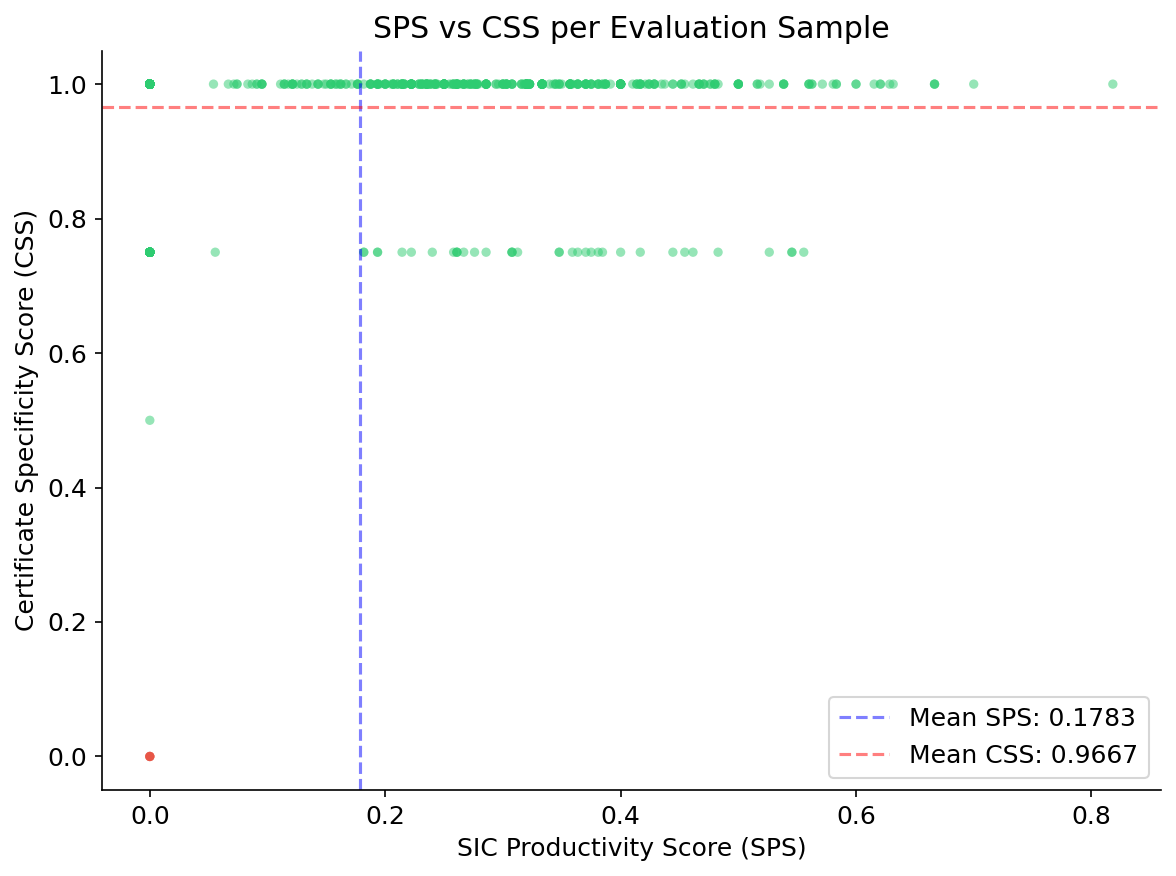}}
    \caption{
      \textbf{SPS vs.\ CSS per Evaluation Sample ($n=735$).}
      Dashed lines mark the means (SPS~=~0.1783, blue; CSS~=~0.9667, red).
      The dense band at CSS~=~1.0 across the full SPS range shows that
      concept specificity is robustly satisfied regardless of retrieval-utility
      performance.
      Red dots (CSS~=~0) are the four invalid JSON completions.
    }
    \label{fig:scatter}
  \end{center}
  \vskip -0.2in
\end{figure}

\paragraph{SPS--CSS Scatter.}
Figure~\ref{fig:scatter} plots SPS against CSS for each evaluation sample.
The vast majority of points cluster at CSS~=~1.0 across the full SPS range,
confirming that specificity is near-universally satisfied regardless of
retrieval-utility performance.
The handful of low-CSS outliers (CSS~$\leq$~0.75) correspond to rare cases
where fewer than three concepts were generated; nearly all have non-zero SPS.
The four red points (CSS~=~0, SPS~=~0) are the invalid JSON outputs.

\begin{figure*}[t]
  \vskip 0.2in
  \begin{center}
    \centerline{\includegraphics[width=\textwidth]{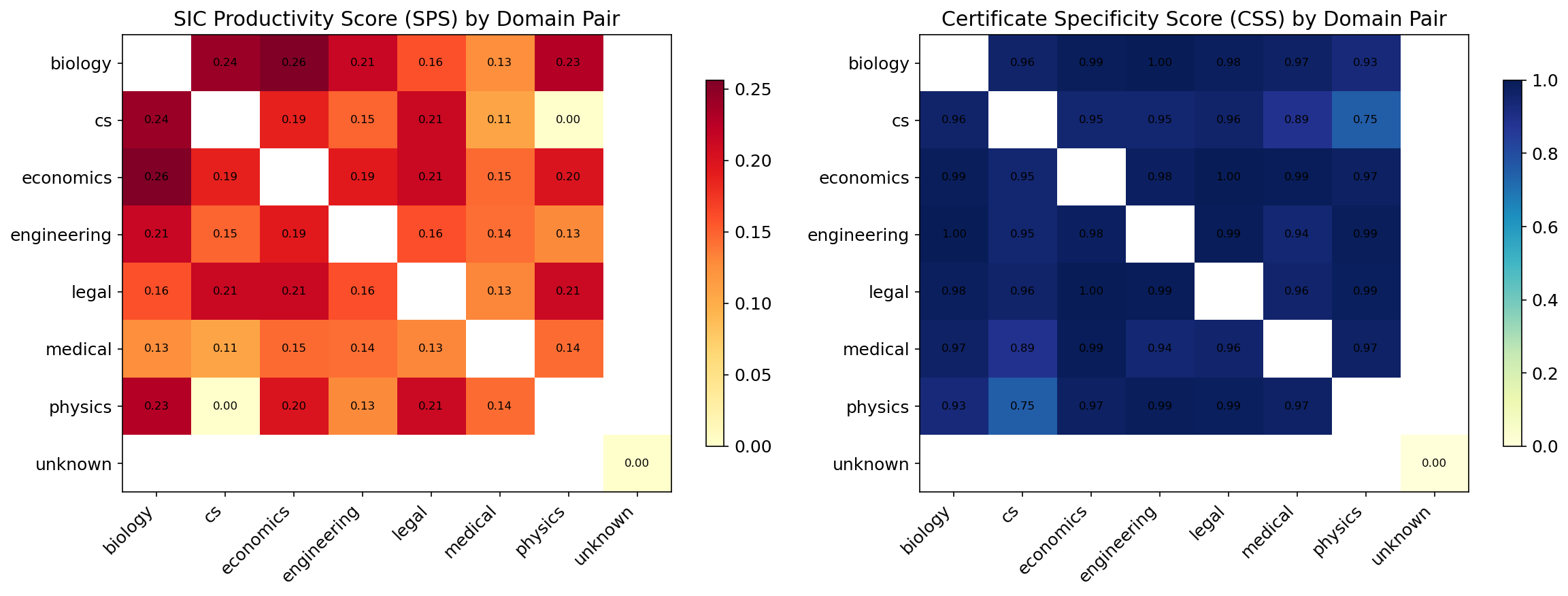}}
    \caption{
      \textbf{SPS and CSS by Domain Pair.}
      \emph{Left}: Mean SIC Productivity Score (ROUGE-L retrieval utility)
      across all 21 domain pairs.  Biology--economics (0.26) and
      biology--physics (0.23) achieve the highest SPS; the cs--physics cell
      (0.00) is an artefact of a boundary split.
      \emph{Right}: Mean Certificate Specificity Score across the same pairs.
      CSS is uniformly high ($\geq$~0.75) everywhere except the cs--physics
      boundary cell, confirming robust concept enumeration across all
      domain combinations.  The \emph{unknown} row/column represents
      domain-pair entries that fell outside the seven primary buckets.
    }
    \label{fig:heatmap}
  \end{center}
  \vskip -0.2in
\end{figure*}

\paragraph{Domain-Pair Breakdown.}
Figure~\ref{fig:heatmap} shows heatmaps of mean SPS (left) and CSS (right)
across all 21 domain pairs.
CSS is uniformly high ($\geq$~0.75) across nearly all pairs.
SPS shows more variation: biology$\times$economics (0.26) and
biology$\times$physics (0.23) score highest; the cs$\times$physics cell
(0.00) is an artefact of a boundary split in the evaluation dataset.
Medical-involving pairs consistently score lower on SPS, reflecting the
high specificity of clinical terminology that is harder to match via ROUGE-L.

\subsection{Ablation: Base vs.\ SIC-Tuned Model}
\label{sec:ablation}

We compare the SIC-tuned model against the base model (LoRA adapters disabled)
on 100 held-out UU questions scored by ROUGE-L against concatenated reference
concepts.

\begin{figure*}[t]
  \vskip 0.2in
  \begin{center}
    \centerline{\includegraphics[width=\textwidth]{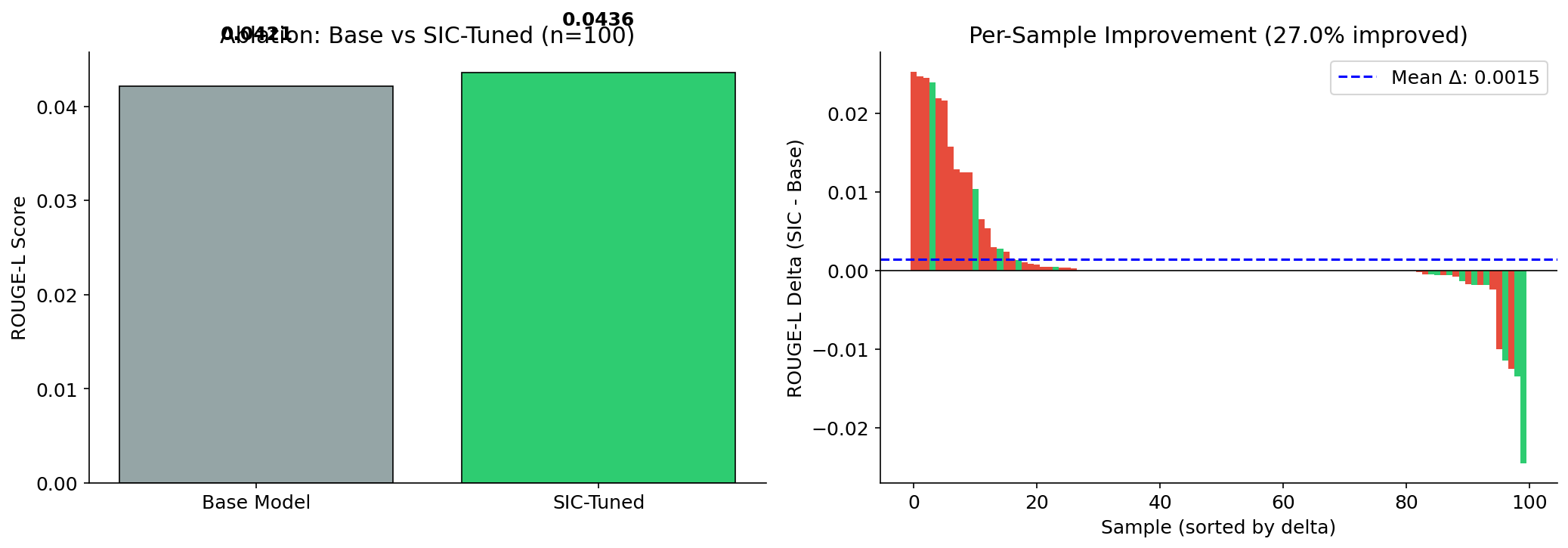}}
    \caption{
      \textbf{Ablation: Base vs.\ SIC-Tuned ($n=100$).}
      \emph{Left}: Mean ROUGE-L scores; the SIC-tuned model (0.0436)
      outperforms the base (0.0421), a 3.6\% relative improvement.
      \emph{Right}: Per-sample ROUGE-L delta (SIC~$-$~Base), sorted
      descending.  Green bars indicate improvements (27\% of samples;
      peak~$+$0.025); the dashed blue line marks the mean delta of
      $+$0.0015.  Regressions are shallower in magnitude than improvements,
      indicating an asymmetric gain profile favourable to the SIC-tuned model.
    }
    \label{fig:ablation}
  \end{center}
  \vskip -0.2in
\end{figure*}

\begin{table}[t]
  \caption{Ablation: Base vs.\ SIC-Tuned ($n=100$, ROUGE-L).}
  \label{tab:ablation}
  \begin{center}
    \begin{small}
      \begin{sc}
        \begin{tabular}{lccc}
          \toprule
          Model          & ROUGE-L & $\Delta$  & \% Improved \\
          \midrule
          Base Qwen3-14B & 0.0421  & ---        & ---         \\
          SIC-Tuned      & 0.0436  & $+$0.0015  & 27.0\%      \\
          \bottomrule
        \end{tabular}
      \end{sc}
    \end{small}
  \end{center}
  \vskip -0.1in
\end{table}

Table~\ref{tab:ablation} and Figure~\ref{fig:ablation} confirm a 3.6\%
relative ROUGE-L improvement.
The modest absolute gain (0.0015) is expected: ROUGE-L measures lexical overlap
with a reference not directly optimized during training, and the base model
already produces some relevant concept terms.
Figure~\ref{fig:ablation} (right) shows that positive deltas (up to $+$0.025)
are sharper than negative ones, confirming that SIC fine-tuning is broadly
beneficial across diverse cross-domain question types.

\section{Discussion}

\paragraph{SIC as a Pre-Retrieval Planner.}
The most direct downstream application is as a \emph{pre-retrieval planner}
in a RAG pipeline.  A system first queries the SIC model to obtain a
\texttt{retrieval\_query} calibrated to the knowledge gap; the
\texttt{required\_concepts} list can additionally filter or re-rank retrieved
documents.

\paragraph{Limitations of Paraphrase Divergence.}
The overall probe accuracy of 44.9\% leaves room for improvement.
The KK/KU boundary is particularly ambiguous: hard MMLU questions may elicit
divergent responses similar to UU questions.
Future work could use learned paraphrasers or combine divergence features with
token-level confidence signals.

\paragraph{Scalability.}
The stitching pipeline requires a capable base model and 448 GPU-minutes for
7,347 samples.  Once the dataset is built, fine-tuning a smaller model
(7B or 3B) is straightforward, making the pipeline applicable to
resource-limited settings.

\paragraph{Reward Design.}
The 0.5/0.3/0.2 weighting reflects our assessment that retrieval utility is
the most operationally important property.  Systematic reward-shaping ablations
are left for future work.

\section{Limitations and Future Work}
 
While Structured Ignorance Certificates demonstrate measurable gains in
epistemic structuring, several limitations remain.
First, the paraphrase-divergence probe achieves only 44.9\% three-way
accuracy, leaving the KK/KU boundary poorly resolved; future work should
explore richer probe features such as token-level entropy, hidden-state
geometry, or ensemble-based divergence estimators.
Second, the UU dataset is constructed from seven pre-defined domain buckets,
which may not cover the full breadth of real-world cross-domain knowledge
gaps---particularly in emerging interdisciplinary fields such as computational
biology or AI policy; expanding the domain taxonomy and sourcing from more
diverse corpora (e.g., ArXiv cross-listings, Wikipedia disambiguation pages)
would improve coverage.
Third, the GRPO reward is measured by ROUGE-L against concatenated concept
strings, a lexical proxy that does not capture semantic equivalence; replacing
it with an embedding-based reward or a learned reward model trained on
human-judged SIC quality would likely yield stronger alignment between the
training signal and downstream utility.
Fourth, our experiments are limited to a single 14B-parameter model family
(Qwen3); replicating the pipeline on models of different scales (3B, 7B, 70B)
and architectures (Llama, Mistral, Gemma) would establish whether SIC training
generalises across the LLM landscape.
Finally, the end-to-end integration of SIC outputs into a live RAG pipeline
remains untested; a promising direction is to use the \texttt{retrieval\_query}
field as a first-stage query rewriter and the \texttt{required\_concepts} list
as a re-ranking filter, evaluating downstream answer quality on open-domain
question answering benchmarks such as Natural Questions and TriviaQA \cite{sahoo2026calibrationcollapsesycophancyfinetuning}.
 
\section{Conclusion}
 
We have presented \textbf{Structured Ignorance Certificates} (SICs), a
framework for training language models to produce structured, actionable
epistemic metadata about questions they cannot answer.
Our pipeline combines (i) a scalable cross-domain UU question synthesis
procedure, (ii) a paraphrase-divergence probe for KK/KU/UU classification,
(iii) a composite reward scheme, and (iv) GRPO-based fine-tuning of a
14B-parameter model.
The fine-tuned model achieves 99.46\% JSON validity, mean CSS of 0.967, and
a 3.6\% relative ROUGE-L improvement over the base model.
We hope the SIC framework provides a foundation for LLM-based systems that
degrade gracefully at knowledge boundaries rather than confabulating
confidently.

\section*{Impact Statement}
This paper advances calibration and epistemic reliability of large language
models.  By training models to articulate what they do not know, we aim to
reduce harmful confabulation in high-stakes domains such as medical and legal
question answering.  No personally identifiable information was used; all
source datasets carry permissive licences.

\bibliography{example_paper}
\bibliographystyle{icml2026}

\newpage
\appendix
\onecolumn

\section{Prompt Details}
\label{app:prompts}

\subsection{UU Stitching Prompt}
\label{app:uu_prompt}

The UU stitching prompt uses the \texttt{/no\_think} directive (supported by
Qwen3 models) to suppress chain-of-thought reasoning tokens during dataset
construction and employs a partial JSON prefix in the assistant turn to nudge
generation toward valid output.

\begin{verbatim}
<|im_start|>system
You are a JSON-only output machine. /no_think
<|im_end|>
<|im_start|>user
You are an expert at constructing difficult cross-domain
questions. Given two questions from different domains,
create ONE new question that GENUINELY requires combining
knowledge from BOTH domains.

Domain A ({domain_a}): {q_a}
Domain B ({domain_b}): {q_b}

Output ONLY valid JSON:
{
  "stitched_question": "...",
  "domain_a": "{domain_a}",
  "domain_b": "{domain_b}",
  "required_concepts_a": ["c1", "c2"],
  "required_concepts_b": ["c1", "c2"],
  "uu_confidence": 0.8
}
<|im_end|>
<|im_start|>assistant
{
\end{verbatim}

Any output not beginning with \texttt{\{} is repaired by prepending the missing
opening brace during post-processing.

\subsection{SIC System Prompt}
\label{app:sic_prompt}

\begin{verbatim}
You are an epistemic reasoner. When given a question,
your job is to generate a Structured Ignorance Certificate
(SIC) - a precise diagnosis of WHY the question exceeds
your knowledge boundary and WHAT information would be
needed to answer it. A good SIC:
1. Identifies the missing domain intersection
2. Names specific missing concepts (not vague)
3. Suggests a productive retrieval query that would
   unlock the answer
4. Avoids generic hedging like "I'm not sure"

Output format (JSON only):
{
  "missing_intersection": "...",
  "required_concepts": ["...", "..."],
  "retrieval_query": "...",
  "confidence_of_ignorance": 0.0
}
\end{verbatim}

\section{Implementation Details}
\label{app:impl}

\subsection{Full Hyperparameter Table}

Table~\ref{tab:hparams} lists all hyperparameters used in the experiments.

\begin{table}[h]
  \caption{Full Hyperparameter Configuration (\texttt{SICConfig}).}
  \label{tab:hparams}
  \begin{center}
    \begin{small}
      \begin{sc}
        \begin{tabular}{lc}
          \toprule
          Parameter              & Value \\
          \midrule
          \multicolumn{2}{l}{\textit{Base Model}} \\
          Model                  & Qwen/Qwen3-14B \\
          Quantization           & 4-bit NF4 \\
          Compute dtype          & bfloat16 \\
          Double quant           & True \\
          Attention              & Flash Attention 2 \\
          \midrule
          \multicolumn{2}{l}{\textit{LoRA}} \\
          Rank $r$               & 16 \\
          $\alpha$               & 32 \\
          Dropout                & 0.05 \\
          Target modules         & q/k/v/o/gate/up/down proj \\
          Trainable params       & 64.2M / 14.8B (0.43\%) \\
          \midrule
          \multicolumn{2}{l}{\textit{GRPO Training}} \\
          Max steps              & 500 \\
          Batch size (device)    & 2 \\
          Grad.\ accum.\ steps   & 4 \\
          Effective batch        & 8 \\
          Learning rate          & $2{\times}10^{-5}$ \\
          Weight decay           & 0.01 \\
          Max grad norm          & 1.0 \\
          Warmup ratio           & 0.05 \\
          Max prompt length      & 512 \\
          Max completion length  & 128 \\
          Rollouts per prompt    & 2 \\
          KL coefficient $\beta$ & 0.04 \\
          Clip $\epsilon$        & 0.2 \\
          GPU                    & NVIDIA A100-SXM4-40GB \\
          VRAM                   & 10.23 GB \\
          Wall-clock time        & $\approx$9 h \\
          \midrule
          \multicolumn{2}{l}{\textit{Dataset}} \\
          SE train size          & 40,000 \\
          UU valid samples       & 7,347 \\
          Train / eval split     & 6,612 / 735 \\
          \midrule
          \multicolumn{2}{l}{\textit{Probe}} \\
          Embed model            & all-MiniLM-L6-v2 \\
          Num.\ paraphrases      & 5 \\
          Classifier             & Logistic Reg.\ (balanced) \\
          KK / KU / UU size      & 300 each \\
          \bottomrule
        \end{tabular}
      \end{sc}
    \end{small}
  \end{center}
\end{table}

\subsection{Multi-Stage JSON Parser}
\label{app:parser}

The \texttt{extract\_json\_from\_response} function implements a six-step
recovery pipeline:

\begin{enumerate}
  \item \textbf{Preprocessing}: Strip \texttt{</think>} tags, Unicode control
    characters, and placeholder strings (e.g., \texttt{"c1"}, \texttt{"c2"}).
  \item \textbf{Direct parse}: Attempt \texttt{json.loads} on the cleaned text.
  \item \textbf{Brace-scan}: Walk character-by-character tracking brace depth;
    collect all complete \texttt{\{...\}} blocks containing
    \texttt{"stitched\_question"}; select the candidate with the longest
    \texttt{stitched\_question} value.
  \item \textbf{Regex fallback}: Extract \texttt{stitched\_question},
    \texttt{domain\_a}, and \texttt{domain\_b} via named-group regex and
    construct a minimal valid object.
  \item \textbf{Full-text parse}: Attempt \texttt{json.loads} on the entire
    unprocessed response.
  \item \textbf{Prefix repair}: If the cleaned text does not start with
    \texttt{\{}, prepend it and retry \texttt{json.loads}.
\end{enumerate}

\subsection{Evaluation Generation Protocol}
\label{app:eval_gen}

A key implementation subtlety arises from \emph{left-padding} used during
batched tokenization.  Because the tokenizer is initialized with
\texttt{padding\_side="left"}, input IDs for a batch of different-length
prompts are padded on the left.
Slicing the generated output at the non-pad token count of the input is
therefore incorrect.
The correct approach records \texttt{input\_len = inputs.input\_ids.shape[1]}
\emph{before} calling \texttt{model.generate}, then slices each output
sequence at position \texttt{input\_len} to recover only the newly generated
tokens.

\section{Training Dynamics}
\label{app:training}

Table~\ref{tab:training_loss} reports training and validation loss at the four
evaluation checkpoints during GRPO training.

\begin{table}[h]
  \caption{GRPO Training Loss Curve.}
  \label{tab:training_loss}
  \begin{center}
    \begin{small}
      \begin{sc}
        \begin{tabular}{ccc}
          \toprule
          Step & Train Loss & Val.\ Loss \\
          \midrule
          100  & 0.01435    & 0.01504    \\
          200  & 0.01549    & 0.01955    \\
          300  & 0.00370    & 0.01309    \\
          400  & 0.01744    & 0.01702    \\
          500  & 0.00967    & 0.01732    \\
          \bottomrule
        \end{tabular}
      \end{sc}
    \end{small}
  \end{center}
\end{table}

The loss curve exhibits the noisy, non-monotone behaviour typical of
policy-gradient training with small group sizes ($G=2$).
Step 300 shows a notable dip in both train and validation loss, suggesting a
phase transition where the policy consolidates the JSON format constraint.
Final training loss (0.00967) is lower than validation loss (0.01732),
indicating mild overfitting; future work should explore larger evaluation sets
and early stopping.

\section{Qualitative SIC Examples}
\label{app:examples}

Table~\ref{tab:examples} shows three representative SICs produced by the
fine-tuned model on held-out test questions, illustrating the range of
cross-domain coverage.

\begin{table*}[h]
  \caption{Qualitative SIC Examples from the Fine-Tuned Model.}
  \label{tab:examples}
  \begin{center}
    \begin{small}
      \begin{tabular}{p{4.5cm} p{4.5cm} p{4.5cm}}
        \toprule
        \textbf{physics $\times$ legal} &
        \textbf{engineering $\times$ cs} &
        \textbf{biology $\times$ economics} \\
        \midrule
        A satellite operator is legally liable for damages caused by a
        collision in low Earth orbit.  How do Newtonian mechanics and the
        Outer Space Treaty jointly determine fault? &
        When designing a quantum computing algorithm using surface codes,
        how should classical control overhead be minimized given qubit
        connectivity constraints? &
        How does the evolutionary dynamics of antibiotic resistance interact
        with pharmaceutical patent law to affect the long-run supply of
        effective antibiotics? \\
        \midrule
        \textbf{Missing intersection}: Newtonian mechanics and international
        space law in satellite collision liability. &
        \textbf{Missing intersection}: Quantum computing architecture and
        classical control system integration. &
        \textbf{Missing intersection}: Evolutionary biology and pharmaceutical
        economics of antibiotic development. \\
        \midrule
        \textbf{Concepts}: Outer Space Treaty liability; Liability Convention
        due diligence; orbital collision avoidance; legal definitions of fault
        in space operations. &
        \textbf{Concepts}: Surface code topological constraints; logical qubit
        encoding efficiency; classical control overhead; qubit connectivity
        optimization. &
        \textbf{Concepts}: Resistance selection pressure; market exclusivity
        periods; antibiotic stewardship economics; patent cliff dynamics. \\
        \midrule
        \textbf{Retrieval query}: ``Intersection of Newtonian mechanics and
        international space law in determining liability for satellite
        collisions under the Liability Convention and OST'' &
        \textbf{Retrieval query}: ``Optimizing surface code logical qubit
        layouts for minimal physical qubits with fault-tolerant operations
        and classical control efficiency'' &
        \textbf{Retrieval query}: ``Economic incentives for antibiotic
        development under evolutionary resistance pressure and patent
        exclusivity'' \\
        \bottomrule
      \end{tabular}
    \end{small}
  \end{center}
\end{table*}

\section{Domain Keyword Lists}
\label{app:keywords}

Table~\ref{tab:keywords} lists the keyword lists used for domain tagging
(partial; each list contains 12--20 terms in the implementation).

\begin{table}[h]
  \caption{Domain Keywords Used for Bucket Construction (partial list).}
  \label{tab:keywords}
  \begin{center}
    \begin{small}
      \begin{tabular}{ll}
        \toprule
        Domain      & Sample Keywords \\
        \midrule
        Physics     & quantum, relativity, thermodynamics, photon,
                      momentum, gravity, wavelength, electron \\
        Biology     & genome, protein, enzyme, mutation, dna, rna,
                      photosynthesis, membrane, immune \\
        Engineering & circuit, torque, voltage, transistor, bridge,
                      stress, hydraulic, turbine, structural \\
        CS          & algorithm, complexity, neural, compiler, hash,
                      database, recursive, sorting, thread \\
        Economics   & elasticity, arbitrage, equilibrium, inflation,
                      gdp, fiscal, monetary, unemployment \\
        Medical     & (MedQA-USMLE training split; no keywords needed) \\
        Legal       & law, court, contract, liability, statute,
                      jurisdiction, plaintiff, patent, tort \\
        \bottomrule
      \end{tabular}
    \end{small}
  \end{center}
\end{table}

\end{document}